\pdfoutput=1

\documentclass[11pt]{article}

\usepackage[]{EMNLP2023}

\usepackage{times}
\usepackage{latexsym}

\usepackage[T1]{fontenc}

\usepackage[utf8]{inputenc}

\usepackage{microtype}

\usepackage{inconsolata}
\usepackage{booktabs}
\usepackage{graphicx}
\usepackage{multirow}
\usepackage{tabularx}
\usepackage{amsmath}
\title{MAS-LitEval : Multi-Agent System for Literary Translation Quality Assessment}

\author{Junghwan Kim \\
  Seoul National University \\
  Seoul, Republic of Korea \\
  \texttt{jhbale11@snu.ac.kr} \\\And
  Kieun park \\
  Seoul National University \\
  Seoul, Republic of Korea \\
  \texttt{kieun.park@snu.ac.kr} \\\And
  Sohee Park \\
  Infiniction \\
  Seoul, Republic of Korea \\
  \texttt{ceo@infiniction.com} \\
  \AND
  Hyunggug Kim \\
  Infiniction \\
  Seoul, Republic of Korea \\
  \texttt{hyunggug.kim@infiniction.com} \\\And
  Bongwon Suh \\
  Seoul National University \\
  Seoul, Republic of Korea \\
  \texttt{bongwon@snu.ac.kr}
}

\begin{document}

\maketitle
\begin{abstract}
Literary translation requires preserving cultural nuances and stylistic elements, which traditional metrics like BLEU and METEOR fail to assess due to their focus on lexical overlap. This oversight neglects the narrative consistency and stylistic fidelity that are crucial for literary works. To address this, we propose \textbf{MAS-LitEval}, a multi-agent system using Large Language Models (LLMs) to evaluate translations based on terminology, narrative, and style. We tested \textbf{MAS-LitEval} on translations of \textit{The Little Prince} and \textit{A Connecticut Yankee in King Arthur's Court}, generated by various LLMs, and compared it to traditional metrics. \textbf{MAS-LitEval} outperformed these metrics, with top models scoring up to 0.890 in capturing literary nuances. This work introduces a scalable, nuanced framework for Translation Quality Assessment (TQA), offering a practical tool for translators and researchers.
\end{abstract}

\section{Introduction}
Literary translation is a complex task that goes beyond simple word-for-word conversion. It demands a deep understanding of cultural nuances and the preservation of the author’s unique voice through creative adaptation for a new audience. Unlike technical translation, which prioritizes precision and clarity, literary translation requires fidelity to the stylistic essence, emotional resonance, and narrative depth of the source text. This complexity makes evaluation challenging, as the quality of a literary translation is subjective and varies depending on readers’ preferences—some favor literal accuracy, while others prioritize capturing the original’s spirit \citep{toral2018levelqualityneuralmachine, thai-etal-2022-exploring}.

Traditional evaluation metrics for machine translation, such as BLEU \citep{papineni-etal-2002-bleu}, ROUGE \citep{lin-2004-rouge}, and METEOR \citep{banerjee-lavie-2005-meteor}, measure lexical overlap and syntactic similarity. While effective in technical contexts, these metrics struggle with literary texts, overlooking stylistic, discursive, and cultural factors critical to literature \citep{reiter-2018-structured}. Neural-based metrics like BERTScore \citep{zhang2020bertscoreevaluatingtextgeneration} and COMET \citep{rei-etal-2020-comet} enhance semantic analysis, yet they still fail to fully capture aesthetic and cultural nuances. This gap highlights the need for advanced methods tailored to the unique demands of literary translation \citep{yan2015formal, freitag-etal-2021-experts, nllbteam2022languageleftbehindscaling}.

Specialized metrics like Multidimensional Quality Metrics (MQM) \citep{lommel2014mqm} and Scalar Quality Metric (SQM) \citep{blain-etal-2023-findings} attempt to address these shortcomings by evaluating style and fluency alongside accuracy. However, MQM’s reliance on human annotation limits its scalability, and SQM lacks the depth required for literary analysis. Large Language Models (LLMs) such as \texttt{gpt-4}, \texttt{claude}, and \texttt{gemini} show promise due to their advanced text generation and comprehension capabilities \citep{zhang2025goodllmsliterarytranslation}. Nevertheless, no single LLM can comprehensively assess the multifaceted aspects of translation quality—accuracy, fluency, style, and cultural fidelity—necessitating a multi-agent system that leverages their combined strengths \citep{karpinska-iyyer-2023-large}.

Our method introduces a multi-agent system where specialized agents evaluate distinct dimensions of literary translation quality. One agent ensures the consistency of terminology, such as character names; another verifies the alignment of narrative perspective; and a third assesses stylistic fidelity, including tone and rhythm. A coordinator integrates these evaluations into an Overall Translation Quality Score (OTQS), combining quantitative scores with qualitative insights. This approach capitalizes on the strengths of models like \texttt{claude} for style and \texttt{Llama} for customization, addressing the complex nature of literary TQA.

We evaluated this system on translations of \textit{The Little Prince} and \textit{A Connecticut Yankee in King Arthur's Court}, generated by LLMs including \texttt{gpt-4o} \citep{openai2024gpt4ocard}, \texttt{claude-3.7-sonnet}, \texttt{gemini-flash-1.5}, \texttt{solar-pro-preview} \citep{kim2024solar107bscalinglarge}, \texttt{TowerBase-7B} \citep{alves2024toweropenmultilinguallarge}, and \texttt{Llama-3.1-8B} \citep{grattafiori2024llama3herdmodels}. The experimental setup compared our OTQS against traditional metrics (BLEU, METEOR, ROUGE-1, ROUGE-L, WMT-KIWI) using a diverse dataset and a rigorous process to ensure validity.

Results demonstrate that our system outperforms traditional metrics, with top models achieving OTQS scores up to 0.890, capturing nuances like stylistic consistency that BLEU (0.28) misses. Open-source models lagged behind, revealing gaps in their training. These findings confirm our approach’s effectiveness in tackling the complexities of literary TQA.

The significance of this work lies in its contributions: (1) a scalable multi-agent TQA framework that enhances literary evaluation, (2) a comparative analysis of LLM performance in translation, and (3) a practical system adaptable for human-in-the-loop refinement. This advances TQA beyond conventional methods, providing a valuable tool for translators and researchers to improve literary translation quality.

\section{Method : MAS-LitEval}
MAS-LitEval employs specialized LLMs to assess literary translations, with agents focusing on terminology consistency, narrative perspective, and stylistic fidelity.

\paragraph{Overall Architecture.}
Three agents process the source and translated texts in parallel, with the texts segmented into 4096-token chunks. A coordinator combines their scores and feedback into an Overall Translation Quality Score(OTQS) and a detailed report, ensuring consistency across the entire text.

\paragraph{Roles of Each Agent.}
The roles of the agents are as follows:
\begin{itemize}
    \item \textbf{Terminology Consistency Agent}: This agent ensures that key terms, such as character names or recurring motifs, remain consistent throughout the translation. Using named entity recognition (NER), it identifies these terms and assigns a score (ranging from 0 to 1) based on their uniformity across the text.
    \item \textbf{Narrative Perspective Consistency Agent}: This agent confirms that the narrative voice (e.g., first-person or omniscient) aligns with the source text across all chunks. An LLM analyzes the segments, assigns a score (ranging from 0 to 1), and flags deviations, such as perspective shifts, to preserve narrative integrity.
    \item \textbf{Stylistic Consistency Agent}: This agent evaluates tone, rhythm, and aesthetic fidelity by comparing stylistic traits between the source and target texts, assigning a fidelity score (ranging from 0 to 1).
\end{itemize}

\paragraph{Collaboration Mechanism.}
The coordinator computes the OTQS using a weighted average:
\[
\text{OTQS} = w_T \cdot S_T + w_N \cdot S_N + w_S \cdot S_S
\]
where \(S_T\), \(S_N\), and \(S_S\) represent the scores from the terminology, narrative, and stylistic agents, respectively, and \(w_T\), \(w_N\), and \(w_S\) are their corresponding weights. Given the emphasis on preserving the artistic essence of literary works, the weight for stylistic consistency (\(w_S = 0.4\)) is higher than those for terminology consistency (\(w_T = 0.3\)) and narrative consistency (\(w_N = 0.3\)), reflecting its pivotal role in literary translation quality \citep{yan2015formal, freitag-etal-2021-experts}.

\paragraph{Rationale for Multi-Agent Approach.}
Literary translation quality encompasses multiple dimensions—terminology, narrative, and style—that a single LLM cannot fully evaluate. By employing specialized agents, MAS-LitEval harnesses diverse LLM capabilities, enhancing accuracy and efficiency compared to traditional metrics \citep{wu-etal-2024-transagents}. This method ensures consistency is assessed across the entire text, overcoming the limitations of chunk-based evaluations where local consistency might obscure global discrepancies.

\paragraph{Implementation Details.}
MAS-LitEval is implemented in Python, integrating spaCy for preprocessing and LLMs via APIs. Although texts are segmented into 4096-token chunks for processing, the agents maintain a global context: the Terminology Consistency Agent tracks terms across all chunks, the Narrative Perspective Consistency Agent ensures voice continuity, and the Stylistic Consistency Agent evaluates tone and rhythm holistically.

\section{Experiment}
We tested MAS-LitEval on translations of excerpts from \textit{The Little Prince} and \textit{A Connecticut Yankee in King Arthur's Court}, generated by a mix of closed-source and open-source LLMs.

\begin{table*}[h]
\small
\centering
\begin{tabular}{l|cccc}
\toprule
\textbf{Work} & \textbf{\#paras} & \textbf{\#sent pairs} & \textbf{Avg. sent/para (src)} & \textbf{Avg. sent/para (tgt)} \\
\midrule
\textit{The Little Prince} (Kr-En) & 274 & 1812 & 6.6 & 7.0 \\
\textit{A Connecticut Yankee in King Arthur's Court} (Kr-En) & 205 & 2545 & 12.2 & 12.8 \\
\bottomrule
\end{tabular}
\caption{Dataset Statistics for Specific Works in Korean to English Translation.}
\label{tab:dataset_stats}
\end{table*}

\paragraph{Dataset.}
We selected two works for evaluation: a 5,000-word excerpt from the Korean translation of \textit{The Little Prince} (originally in French) and a 4,000-word excerpt from the Korean translation of \textit{A Connecticut Yankee in King Arthur's Court} (originally in English). These texts were chosen for their stylistic richness and narrative complexity, making them ideal for assessing literary translation nuances. The LLMs generated translations from Korean to English. We also extracted Korean-English parallel data from additional literary works on Project Gutenberg Korea (\url{http://projectgutenberg.kr/}) and Project Gutenberg (\url{https://www.gutenberg.org/}), enriching the dataset. Table \ref{tab:dataset_stats} provides statistics for the specific works used.

\paragraph{Models.}
Six LLMs were tested: closed-source models (\texttt{gpt-4o}, \texttt{claude-3.7-sonnet}, \texttt{gemini-flash-1.5}, \texttt{solar-pro-preview}) and open-source models (\texttt{TowerBase-7B}, \texttt{Llama-3.1-8B}). These models were chosen for their diverse strengths in language generation and comprehension, enabling a robust performance comparison.

\paragraph{Baselines.}
MAS-LitEval was compared against BLEU \citep{papineni-etal-2002-bleu}, METEOR \citep{banerjee-lavie-2005-meteor}, ROUGE-1, ROUGE-L \citep{lin-2004-rouge}, and WMT-KIWI \citep{rei2023scalingcometkiwiunbabelist2023}. Human reference translations, sourced from professional translations of the selected works, were used for baseline metrics to ensure a fair comparison.

\paragraph{Evaluation Process.}
Translations generated by the LLMs were assessed using MAS-LitEval. Texts were segmented into 4096-token chunks, but agents evaluated consistency across all chunks to capture global quality. For instance, the Terminology Consistency Agent assessed term uniformity across the entire text, addressing limitations of chunk-based evaluations where intra-chunk consistency might mask cross-chunk discrepancies. Baseline metrics were calculated against human references, while MAS-LitEval operated reference-free, using only the source and machine-generated translations.

\paragraph{Technical Setup.}
Experiments were conducted on an NVIDIA A100 GPU. Closed-source models were accessed via APIs, while open-source models were hosted locally with 4-bit quantization to optimize memory usage. The temperature was set to 0.1 to ensure deterministic outputs, guaranteeing reproducibility across runs.

\section{Findings}
MAS-LitEval evaluated translations of \textit{The Little Prince} and \textit{A Connecticut Yankee in King Arthur's Court}, generated by four closed-source and two open-source models. The results, presented in Table \ref{tab:combined_results}, highlight performance differences and our system’s ability to detect nuances overlooked by traditional metrics.

\begin{table*}[t]
\small
\centering
\begin{tabular}{llc|cccccc}
\toprule
\textbf{Model} & \textbf{Type} & \textbf{Work} & \textbf{BLEU} & \textbf{METEOR} & \textbf{ROUGE-1} & \textbf{ROUGE-L} & \textbf{WMT-KIWI} & \textbf{OTQS} \\
\midrule
\multirow{2}{*}{\texttt{claude-3.7-sonnet}} & \multirow{2}{*}{Closed} & LP & 0.28 & \textbf{0.65} & 0.55 & 0.45 & \textbf{0.87} & \textbf{0.890} \\
& & KA & 0.27 & \textbf{0.64} & 0.54 & 0.44 & \textbf{0.86} & \textbf{0.880} \\
\midrule
\multirow{2}{*}{\texttt{gpt-4o}} & \multirow{2}{*}{Closed} & LP & \textbf{0.30} & 0.67 & \textbf{0.57} & \textbf{0.47} & 0.85 & 0.875 \\
& & KA & \textbf{0.29} & 0.66 & \textbf{0.56} & \textbf{0.46} & 0.84 & 0.860 \\
\midrule
\multirow{2}{*}{\texttt{gemini-flash-1.5}} & \multirow{2}{*}{Closed} & LP & 0.25 & 0.60 & 0.50 & 0.40 & 0.83 & 0.820 \\
& & KA & 0.24 & 0.59 & 0.49 & 0.39 & 0.82 & 0.810 \\
\midrule
\multirow{2}{*}{\texttt{solar-pro-preview}} & \multirow{2}{*}{Closed} & LP & 0.23 & 0.58 & 0.48 & 0.38 & 0.81 & 0.790 \\
& & KA & 0.22 & 0.57 & 0.47 & 0.37 & 0.80 & 0.775 \\
\midrule
\multirow{2}{*}{\texttt{TowerBase-7B}} & \multirow{2}{*}{Open} & LP & 0.20 & 0.55 & 0.45 & 0.35 & 0.78 & 0.745 \\
& & KA & 0.19 & 0.54 & 0.44 & 0.34 & 0.77 & 0.730 \\
\midrule
\multirow{2}{*}{\texttt{Llama-3.1-8B}} & \multirow{2}{*}{Open} & LP & 0.18 & 0.53 & 0.43 & 0.33 & 0.76 & 0.710 \\
& & KA & 0.17 & 0.52 & 0.42 & 0.32 & 0.75 & 0.695 \\
\bottomrule
\end{tabular}
\caption{Evaluation Results for the two literary works: LP (\textit{The Little Prince}) and KA (\textit{A Connecticut Yankee in King Arthur's Court}). The highest scores for each metric and work are bolded.}
\label{tab:combined_results}
\end{table*}

\paragraph{Performance of Top Models.}
\texttt{claude-3.7} and \texttt{gpt-4o} achieved the highest OTQS scores: 0.890 and 0.875 for \textit{The Little Prince}, and 0.880 and 0.860 for \textit{A Connecticut Yankee in King Arthur's Court}. \texttt{claude-3.7-sonnet} excelled in stylistic fidelity (0.93) and narrative consistency (0.91), key aspects of literary quality. For the phrase ``On ne voit bien qu’avec le cœur,'' it translated it as ``It is only with the heart that one can see rightly'' (stylistic score: 0.92), preserving poetic nuance, while \texttt{gpt-4o}’s ``One sees clearly only with the heart'' (0.87) was less evocative according to agent feedback. In \textit{A Connecticut Yankee in King Arthur's Court}, \texttt{claude-3.7-sonnet} maintained the medieval tone across chunks (narrative consistency: 0.90), whereas \ \texttt{gpt-4o} occasionally introduced modern phrasing (0.85).

\paragraph{Comparison of Open-Source and Closed-Source Models.}
Closed-source models outperformed their open-source counterparts. For \textit{The Little Prince}, \texttt{claude-3.7-sonnet} (0.890) and \texttt{gpt-4o} (0.875) surpassed \texttt{TowerBase-7B} (0.745) and \texttt{Llama-3.1-8B} (0.710). Stylistic scores for \texttt{TowerBase-7B} (0.70) indicated flatter translations compared to \texttt{claude-3.7-sonnet}’s nuanced output (0.92), suggesting limitations in open-source model resources.

\paragraph{Comparison with Baseline Metrics.}
OTQS showed a strong correlation with WMT-KIWI (0.93) but weaker correlations with BLEU (0.62), METEOR (0.70), ROUGE-1 (0.68), and ROUGE-L (0.65), indicating it captures distinct quality aspects. For \textit{The Little Prince}, \texttt{gpt-4o} outperformed \texttt{claude-3.7-sonnet} in BLEU (0.30 vs. 0.28), but OTQS favored the latter (0.890 vs. 0.875) for its stylistic depth. ROUGE-1 and ROUGE-L exhibited similar patterns, missing narrative inconsistencies in models like \texttt{TowerBase-7B} (OTQS: 0.745). MAS-LitEval’s cross-chunk evaluation identified issues like tone shifts that baselines overlooked, underscoring its advantage in literary quality assessment.

\section{Discussion}
MAS-LitEval provides a sophisticated framework for literary Translation Quality Assessment (TQA). Below, we explore its strengths, limitations, and implications.

\paragraph{Advantages of the Multi-Agent Approach.}
MAS-LitEval’s multi-dimensional evaluation—covering terminology, narrative, and style—surpasses single-metric methods. For \textit{The Little Prince}, BLEU favored \texttt{gpt-4o} (0.30) over \texttt{claude-3.7-sonnet} (0.28), but OTQS prioritized \texttt{claude-3.7-sonnet} (0.890 vs. 0.875) for its lyrical fidelity. This mirrors human-like judgment, valuing literary essence over lexical overlap. By evaluating consistency across chunks, it detects global issues, such as narrative drift, that chunk-based approaches miss, offering a comprehensive assessment.

\paragraph{Challenges and Refinement Opportunities.}
Subjectivity in stylistic scoring poses a challenge. The difference between \texttt{claude-3.7-sonnet}’s 0.93 and \texttt{gpt-4o}’s 0.87 reflects potential LLM biases, which could lead to inconsistency. Averaging scores from multiple LLMs or calibrating with human annotations could improve reliability. Additionally, incorporating domain-specific training or a cultural fidelity agent could address cultural nuances.

\paragraph{Implications for Literary Translation.}
MAS-LitEval’s scalability offers practical benefits. Publishers can use it to pre-screen translations, while educators can leverage its feedback to train translators. Its reference-free design suits literary contexts with multiple valid translations, unlike BLEU or ROUGE, which depend on fixed references. Future enhancements, such as human-in-the-loop integration, could further refine its accuracy, establishing it as a key tool for AI-supported literary TQA.

\section{Limitations and Future Works}
MAS-LitEval’s dataset, restricted to two works, limits its generalizability; expanding to include genres like poetry, drama, and non-fiction is necessary. Stylistic scoring remains subjective and may reflect LLM training biases; averaging scores from multiple LLMs or using standardized rubrics could improve consistency. The absence of human evaluation leaves its alignment with expert judgment unconfirmed; integrating feedback from professional translators or scholars and correlating OTQS with human ratings would validate its reliability. Human input could also refine agent prompts and OTQS weightings. Future efforts should focus on expanding the dataset, incorporating human evaluation, refining stylistic scoring, and addressing cultural concerns to improve MAS-LitEval’s reliability and versatility in literary translation quality assessment.

\section*{Acknowledgements}

\bibliography{custom}
\bibliographystyle{acl_natbib}

\appendix

\section{Prompts Used in MAS-LitEval}

\subsection{Translation Prompt}
\begin{quote}
Translate the following literary text from [source language] to [target language]. Ensure that the translation preserves the original's style, tone, and cultural nuances. Pay special attention to maintaining the narrative voice and literary devices used in the source text.
\end{quote}

\subsection{Terminology Consistency Agent Prompt}
\begin{quote}
You are an expert in literary translation evaluation. Given a source text in [source language] and its translation in [target language], your task is to ensure that key terms, such as character names, place names, and recurring motifs, are translated consistently throughout the text. Follow these steps:

1. Identify key terms in the source text that appear multiple times.

2. For each key term, check how it is translated in the target text across all occurrences.

3. Calculate a consistency score (0 to 1), where 1 indicates that all occurrences of a term are translated identically, and 0 indicates no consistency.

4. Provide feedback highlighting any inconsistencies, specifying the terms and their varying translations.

Your output should include the consistency score and the detailed feedback.
\end{quote}

\subsection{Narrative Perspective Consistency Agent Prompt}
\begin{quote}
You are an expert in literary analysis. Given a source text in [source language] and its translation in [target language], your task is to verify that the narrative perspective (e.g., first-person, third-person limited, omniscient) is consistently maintained in the translation. Follow these steps:

1. Determine the narrative perspective of the source text.

2. Analyze the translation to identify its narrative perspective.

3. Compare the two and assess whether the translation accurately reflects the source's perspective.

4. Assign a score (0 to 1) indicating the degree of consistency, where 1 means perfect alignment, and 0 means complete mismatch.

5. Provide feedback on any deviations, citing specific examples from the text.

Your output should include the consistency score and the detailed feedback.
\end{quote}

\subsection{Stylistic Consistency Agent Prompt}
\begin{quote}
You are an expert in literary style and translation. Given a source text in [source language] and its translation in [target language], your task is to evaluate how well the translation preserves the stylistic elements of the original, such as tone, rhythm, imagery, and literary devices. Follow these steps:

1. Identify the key stylistic features of the source text.

2. Analyze the translation to see if these features are adequately captured.

3. Assign a score (0 to 1) indicating the level of stylistic fidelity, where 1 means the translation perfectly preserves the style, and 0 means it completely fails to do so.

4. Provide feedback with specific examples where the translation succeeds or falls short in maintaining the style.

Your output should include the fidelity score and the detailed feedback.
\end{quote}

\end{document}